\documentclass[11pt,letterpaper]{article}
\usepackage{naaclhlt2016}
\usepackage{times}
\usepackage{latexsym}

\usepackage{hyperref}
\usepackage{breakurl}
\usepackage{todonotes}
\usepackage{amsmath}
\usepackage{breqn}
\usepackage{colortbl}
\usepackage{xcolor}
\usepackage{multirow}
\usepackage{color,soul}

\usepackage{subcaption}

\naaclfinalcopy 
\def\naaclpaperid{167} 

 \title{The Role of Context Types and Dimensionality \\ in Learning Word Embeddings\vspace{0.1in}}

\author{
   Oren Melamud$^{\dag\ast}$
\ \ \ \  David McClosky$^{\ddag\ast}$
\ \ \ \  Siddharth Patwardhan$^\diamondsuit$
\ \ \ \  Mohit Bansal$^{\S}$
\vspace{5pt}
\\ 
$^{\dag}$Computer Science Department, Bar-Ilan University, Ramat-Gan, Israel\\
{\tt melamuo@cs.biu.ac.il}
\vspace{5pt}
\\
$^\ddag$Google, New York, NY, USA\\
\tt{dmcc@google.com}
\vspace{5pt}\\
$^\diamondsuit$IBM Watson, Yorktown Heights, NY, USA\\
\tt{siddharth@us.ibm.com}
\vspace{5pt}\\
$^\S$Toyota Technological Institute at Chicago, Chicago, IL, 60637, USA\\
\tt{mbansal@ttic.edu}
}

\date{}

\makeatletter
\def\@xfootnote[#1]{%
  \protected@xdef\@thefnmark{#1}%
  \@footnotetext}
\makeatother

\begin{document}
\maketitle
\begin{abstract}
We provide the first extensive evaluation of how using different types of context to learn skip-gram word embeddings affects performance
on a wide range of intrinsic and extrinsic NLP tasks.
Our results suggest that while intrinsic tasks tend to exhibit a clear preference to particular types of contexts and higher dimensionality, more careful tuning is required for finding the optimal settings for most of the extrinsic tasks that we considered.
Furthermore, for these extrinsic tasks, we find that once the benefit from increasing the embedding dimensionality is mostly exhausted, simple concatenation of word embeddings, learned with different context types, can yield further performance gains. 
As an additional contribution, we propose a new variant of the skip-gram model that learns word embeddings from weighted contexts of substitute words.

\end{abstract}

\footnote[$\ast$]{Majority of work performed while at IBM Watson.}

\section{Introduction}
\label{introduction}

Word embeddings have become increasingly popular lately, proving to be valuable as a source of features in a broad range of 
NLP tasks with limited supervision \cite{turian-10,collobert2011natural,socher2013recursive,bansal-14}. 
\texttt{word2vec}\footnote{\scriptsize{\url{http://code.google.com/p/word2vec/}}} skip-gram \cite{mikolov-13} and \texttt{GloVe}\footnote{\scriptsize{\url{http://nlp.stanford.edu/projects/glove/}}} \cite{pennington-14} are
among the most widely used word embedding models today. Their success is largely due to an efficient and user-friendly implementation that learns high-quality word embeddings from very large corpora.

Both \texttt{word2vec} and \texttt{GloVe} learn low-dimensional continuous vector representations for words
by considering window-based contexts, i.e., context words within some fixed distance of each side of the target word.
However, the underlying models are equally applicable to different choices of context types. 
For example, \newcite{bansal-14} and \newcite{levy2014dependencybased} showed that using syntactic contexts rather than window contexts in \texttt{word2vec} captures \emph{functional} similarity (as in \emph{lion:cat}) rather than \emph{topical} similarity or \emph{relatedness} (as in \emph{lion:zoo}). 
Further, \newcite{bansal-14} and \newcite{melamud2015simple} showed the benefits of such modified-context embeddings in dependency parsing and lexical substitution tasks.
However, to the best of our knowledge, there has not been an extensive evaluation of the effect of multiple, diverse context types on a wide range of NLP tasks.

Word embeddings are typically evaluated on {\em intrinsic} and {\em extrinsic} tasks. Intrinsic tasks mostly include predicting human judgments of semantic relations between words, e.g., as in WordSim-353 \cite{wordsim353}, while extrinsic tasks include various `real' downstream NLP tasks, such as coreference resolution and sentiment analysis.
Recent works have shown that while intrinsic evaluations are easier to perform, their correlation with results on extrinsic evaluations is not very reliable \cite{schnabel2015evaluation,lingevaluation2015}, stressing the importance of the latter.

In this work, we provide the first extensive evaluation of
word embeddings learned with different types of context,
on a wide range of intrinsic similarity and relatedness tasks, and extrinsic NLP tasks, namely dependency parsing, named entity recognition, coreference resolution, and sentiment analysis.
We employ contexts based of  different word window sizes, syntactic dependencies, and a lesser-known substitute words approach \cite{yatbaz:2012}. 
Finally, we experiment with combinations of the above word embeddings, comparing two approaches: (1)~simple vector concatenation that offers a wider variety of features for a classifier to choose and learn weighted combinations from,
and (2)~dimensionality reduction via either Singular Value Decomposition or Canonical Correlation Analysis, which tries to find a smaller subset of features.

Our results suggest that it is worthwhile to carefully choose the right type of word embeddings for
an extrinsic NLP task, rather than rely on intrinsic benchmark results. Specifically, picking the optimal context type and dimensionality is critical.
Furthermore, once the benefit from increasing the embedding dimensionality is mostly exhausted, concatenation of word embeddings learned with different context types can yield further performance gains.

\section{Word Embedding Context Types}
\label{sec:learning}

\subsection{Learning Corpus}
We use a fixed learning corpus
for a fair comparison of
all embedding types: a concatenation of three large English corpora: (1) English Wikipedia 2015, (2) UMBC web corpus \cite{UMBC}, and (3) English Gigaword (LDC2011T07) newswire corpus \cite{parker2011english}.
Our concatenated corpus is diverse and substantial in size with approximately 10B~words. This allows us to learn high quality embeddings that cover a large vocabulary.
After extracting clean text from these corpora, we used Stanford CoreNLP \cite{corenlp} for sentence splitting, tokenization, part-of-speech tagging and dependency parsing.\footnote{Parses follow the Universal Dependencies formalism and were produced by Stanford CoreNLP, version 3.5.2} Then, all tokens were lowercased, and sentences were shuffled to prevent structured bias. 
When learning word embeddings, we ignored words with corpus frequency lower than 100, yielding a vocabulary of about 500K words.\footnote{Our word embeddings are available at: \scriptsize{\url{www.cs.biu.ac.il/nlp/resources/downloads/embeddings-contexts/}}}

\subsection{Window-based Word Embeddings}
We used \texttt{word2vec}'s skip-gram model with negative sampling \cite{mikolov-13b} to learn window-based word embeddings.
\footnote{We used negative sampling = 5 and iterations = 3 in all of the experiments described in this paper.}
This popular method embeds both target
words and contexts in the same low-dimensional space, where the embeddings of a target and context are pushed closer together the more frequently they co-occur in a learning corpus. Indirectly, this also results in similar embeddings for target words that co-occur with similar contexts. More formally, this method optimizes the following objective function:
\begin{dmath}
\label{eq:w2v_obj}
L = \sum_{(t,c) \in \mathit{PAIRS}} L_{t,c}
\end{dmath}

\begin{dmath}
\label{eq:w2v_obj_pair}
L_{t,c} = \log \sigma (v'_{c} \cdot v_{t}) + \hspace{-20pt} \sum_{neg \in \mathit{NEGS}_{(t,c)}} \hspace{-20pt} \log \sigma (- v'_{neg} \cdot v_{t})
\end{dmath}
\noindent where $v_{t}$ and $v'_{c}$ are the vector representations of target word $t$ and context word $c$. $\mathit{PAIRS}$ is the set of window-based co-occurring target-context pairs considered by the model that depends on the window size, and $\mathit{NEGS}_{(t,c)}$ is a set of randomly sampled context words used with the pair~$(t,c)$.\footnote{For more details refer to \newcite{mikolov-13b}.}

We experimented with window sizes of 1, 5, and 10, and various dimensionalities. We denote a window-based word embedding with window size of $n$ and dimensionality of $m$ with W\emph{n}$^m$. For example, W5$^{300}$ is a word embedding learned using a window size of 5 and dimensionality of 300.

\subsection{Dependency-based Word Embeddings}
We used \texttt{word2vecf}\footnote{\scriptsize{\url{http://bitbucket.org/yoavgo/word2vecf}}} \cite{levy2014dependencybased}, to learn dependency-based word embeddings from the parsed version of our corpus, similar to the approach of \newcite{bansal-14}. \texttt{word2vecf} accepts as its input arbitrary target-context pairs. In the case of dependency-based word embeddings, the context elements are the syntactic contexts of the target word, rather than the words in a window around it. Specifically, following \newcite{levy2014dependencybased}, we first `collapsed' prepositions (as implemented in \texttt{word2vecf}).
Then, for a target word~$t$ with modifiers $m_{1}$,...,$m_{k}$ and head~$h$, we paired the target word with the context elements
$(m_{1},r_{1})$,...,$(m_{k},r_{k})$,$(h,r_{h}^{-1})$, 
where $r$ is the type of the dependency relation between the head and the modifier (e.g., \emph{dobj}, \emph{prep\_of}) and $r^{-1}$ denotes an inverse relation. We denote a dependency-based word embedding with dimensionality of $m$ by DEP$^m$. We note that under this setting \texttt{word2vecf} optimizes the same objective function described in Equation~(\ref{eq:w2v_obj}), with $\mathit{PAIRS}$ now comprising dependency-based pairs instead of window-based ones.

\subsection{Substitute-based Word Embeddings}

Substitute vectors are a recent approach to representing contexts of target words, proposed in \newcite{yatbaz:2012}. Instead of the neighboring words themselves, a substitute vector includes the potential filler words
for the target word slot, weighted according to how `fit' they are to fill the target slot \emph{given} the neighboring words.
For example, the substitute vector representing the context of the word \emph{love} in ``I \underline{love} my job'', could look like: [\emph{quit}~0.5, \emph{love}~0.3, \emph{hate}~0.1, \emph{lost}~0.1].
Substitute-based contexts are generated using a language model and were successfully used in distributional semantics models for part-of-speech induction \cite{yatbaz:2012}, word sense induction \cite{baskaya-EtAl:2013}, functional semantic similarity \cite{melamud:conll:2014} and lexical substitution tasks \cite{melamud:naacl:2015}.

Similar to \newcite{yatbaz:2012}, we consider the words in a substitute vector, as a weighted set of contexts `co-occurring' with the observed target word.
For example, the above substitute vector is considered as the following set of weighted target-context pairs: \{(\emph{love}, \emph{quit}, 0.5), 
 (\emph{love}, \emph{love}, 0.3), (\emph{love}, \emph{hate}, 0.1), (\emph{love}, \emph{lost}, 0.1)\}. 
To learn word embeddings from such weighted target-context pairs, we extended \texttt{word2vecf} by modifying the objective function in Equation (\ref{eq:w2v_obj}) as follows: 
\begin{dmath}
\label{eq:w2v_obj_weighted}
L = \sum_{(t,c) \in \mathit{PAIRS}} \alpha_{t,c} \cdot L_{t,c}
\end{dmath}
\noindent where $\alpha_{t,c}$ is the weight of the target-context pair~($t,c$).
With this simple modification, the effect of target-context pairs on the learned word representations becomes proportional to their weights.

To generate the substitute vectors we followed the methodology in \cite{yatbaz:2012,melamud:naacl:2015}. 
We learned a 4-gram Kneser-Ney language model from our learning corpus using KenLM \cite{kenlm:2013}.
Then, we used FASTSUBS \cite{yuret2012fastsubs} with this language model to efficiently generate substitute vectors, where the weight of each substitute $s$ is the conditional probability $p(s|C)$
for this substitute to fill the target slot given the sentential context $C$. For efficiency, we pruned the substitute vectors to their top-10 substitutes, $s_{1}..s_{10}$, and normalized their probabilities such that $\sum_{i=1..10}{{p}(s_{i}|C)}~=~1$. 
We also generated only up to 20,000 substitute vectors for each target word type.
Finally, we converted each substitute vector into weighted target-substitute pairs and used our extended version of \texttt{word2vecf} to learn the substitute-based word embeddings, denoted SUB$^m$.

\subsection{Qualitative Effect of Context Type}

\begin{table}[t]
\centering
\begin{tabular}{| l  l  l |}
\hline
W10$^{300}$ & DEP$^{300}$ & SUB$^{300}$ \\
\hline
played & play & singing  \\
play  & played & rehearsing  \\
plays  & understudying & performing  \\
professionally  & caddying & composing  \\
player  & plays & running \\
\hline
\end{tabular}
\caption{The top five words closest to target word \emph{playing} in different embedding spaces.
}
\label{tab:top}
\end{table}

To motivate the rest of our work, we first qualitatively inspect the top most-similar words to some target words, using cosine similarity of their respective embeddings. As illustrated in Table~\ref{tab:top}, in embeddings learned with large window contexts, we see both functionally similar words and topically similar words, sometimes with a different part-of-speech. With small windows and dependency contexts, we generally see much fewer topically similar words, which is consistent with previous findings \cite{bansal-14,levy2014dependencybased}. Finally, with substitute-based contexts, there appears to be even a stronger preference for functional similarity, with a tendency to also strictly preserve verb tense.

\section{Word Embedding Combinations}
\label{sec:combinations}
As different choices of context type yield word embeddings with different properties, we hypothesize that combinations of such embeddings could be more informative for some extrinsic tasks. 
We experimented with two alternative approaches to combine different sets of word embeddings: (1) Simple vector concatenation, which is a lossless combination that comes at the cost of increased dimensionality, and (2) SVD and CCA, which are lossy combinations that attempt to capture the most useful information from the different embeddings sets with lower dimensionality. The methods used are described in more detail next.

\subsection{Concatenation}
\label{subsec:concats}

Perhaps the simplest way to combine two different sets of word embeddings (sharing the same vocabulary) is to concatenate their word vectors for every word type. We denote such a combination of word embedding set $A$ with word embedding set $B$ using the symbol (\emph{+}). For example W10+DEP$^{600}$ is the concatenation of W10$^{300}$ with DEP$^{300}$. Naturally, the dimensionality of the concatenated embeddings is the sum of the dimensionalities of the component embeddings. In our experiments, we only ever combine word embeddings of equal dimensionality.

The motivation behind concatenation relates primarily to supervised models in extrinsic tasks. In such settings, we hypothesize that using concatenated word embeddings as input features to a classifier could let it choose and combine (i.e., via learned weights) the most suitable features for the task. Consider a situation where the concatenated embedding W10+DEP$^{600}$ is used to represent the word inputs to a named entity recognition classifier. In this case, the classifier could choose, for instance, to represent entity words mostly with dependency-based embedding features (reflecting functional semantics), and surrounding words with large window-based embedding features (reflecting topical semantics).

\subsection{Singular Value Decomposition}
\label{subsec:svd}
Singular Value Decomposition (SVD) has been shown to be effective in compressing sparse word representations \cite{levy2015improving}. In this work, we use this technique in the same way to reduce the dimensionality of concatenated word embeddings.

\subsection{Canonical Correlation Analysis}
\label{subsec:cca}

Recent work
used Canonical Correlation Analysis (CCA) to 
derive an improved set of word embeddings.
The main idea is that two distinct sets of word embeddings, learned with different types of input data, are considered as multi-views of the same vocabulary. Then, CCA is used to project each onto a lower dimensional space, where correlation between the two is maximized. The correlated information is presumably more reliable.
\newcite{dhillon2011cca} considered their two CCA views as embeddings learned from the left and from the right context of the target words, showing improvements on chunking and named entity recognition.
\newcite{faruqui2014cca} and \newcite{lu2015deep} considered multilingual views,
showing improvements in several intrinsic tasks, such as word and phrase similarity.

Inspired by this prior work, we consider pairs of word embedding sets, learned with different types of context, as different views and correlate them using linear CCA.\footnote{See \newcite{faruqui2014cca}, \newcite{lu2015deep} for details.} 
We use either the SimLex-999 or WordSim-353-R intrinsic benchmark (section \ref{subsec:intrinsic}) to tune the CCA hyperparameters\footnote{These are projection dimensionality and regularization.}
with the Spearmint Bayesian optimization tool\footnote{\url{github.com/JasperSnoek/spearmint}} \cite{spearmint}.
This results in different projections for each of these tuning objectives, where SimLex-999/WordSim-353-R is expected to give some bias towards functional/topical similarity, respectively.

\section{Evaluation}
\label{evaluation}

\subsection{Intrinsic Benchmarks}
\label{subsec:intrinsic}

We employ several commonly used intrinsic benchmarks for assessing how well word embeddings mimic human judgements of semantic similarity of words. The popular \textbf{WordSim-353} dataset \cite{wordsim353} includes 353 word pairs manually annotated with a degree of similarity. For example, \emph{computer}:\emph{keyboard} is annotated with 7.62, indicating a relatively high degree of similarity. 
While \mbox{WordSim-353} does not make a distinction between different `flavors' of similarity, \newcite{agirre:2009} proposed two subsets of this dataset, \textbf{WordSim-353-S} and \textbf{WordSim-353-R}, which focus on functional and topical similarities, respectively.
\textbf{SimLex-999} \cite{hill2014simlex} is a larger word pair similarity dataset with 999 annotated pairs, purposely built to focus on functional similarity.
We evaluate our embeddings on these datasets by computing a score for each pair as the cosine similarity of two word vectors.
The Spearman's correlation\footnote{We used \texttt{spearmanr}, SciPy version 0.15.1.} between the ranking of word pairs induced from the human annotations and that from the embeddings is reported.

The \textbf{TOEFL} task contains 80 synonym selection items, where a synonym of a target word is to be selected out of four possible choices. 
We report the overall accuracy of a system that uses cosine distance between the embeddings of the target word and each of the choices to select the one most similar to the target word as the answer. 

\subsection{Extrinsic Benchmarks}
\label{subsec:extrinsic}
The following four diverse downstream NLP tasks serve as our extrinsic benchmarks.\footnote{Since our goal is to explore performance trends, we mostly experimented with the tasks' development sets.}

\paragraph{1) Dependency Parsing (\textsc{parse})} The Stanford Neural Network Dependency (NNDEP) parser  \cite{chen2014fast} uses 
dense continuous representations of words, parts-of-speech and dependency labels. While it can learn these representations entirely during the training on labeled data, \newcite{chen2014fast} show that initialization with word embeddings, which were pre-trained on unlabeled data, yields improved performance.
Hence, we used our different types of embeddings to initialize the NNDEP parser and compared their performance on a standard Penn Treebank benchmark. We used WSJ sections 2--21 for training and 22 for development. We used predicted tags produced via 20-fold jackknifing on sections 2--21 with the Stanford CoreNLP tagger. 

\paragraph{2) Named Entity Recognition (\textsc{ner})} We used the NER system of \newcite{turian-10}, which allows adding word embedding features (on top of various other features) to a regularized averaged perceptron classifier, and achieves near state-of-the-art results using several off-the-shelf word representations.
We varied the type of word embeddings used as features when training the NER model, to evaluate their effect on NER benchmarks results. Following \newcite{turian-10}, we used the CoNLL-2003 shared task dataset \cite{conll-2003-ner} with 204K/51K train/dev words, as our main benchmark.
We also performed an out-of-domain evaluation, using CoNLL-2003 as the train set and the MUC7 formal run (59K words) as the test set.\footnote{See \newcite{turian-10} for more details on this setting.}

\newcommand{\figsize}{0.43}

\begin{figure*}[t]
\centering
$
\begin{array}{cc}
\includegraphics[width=\figsize\textwidth]{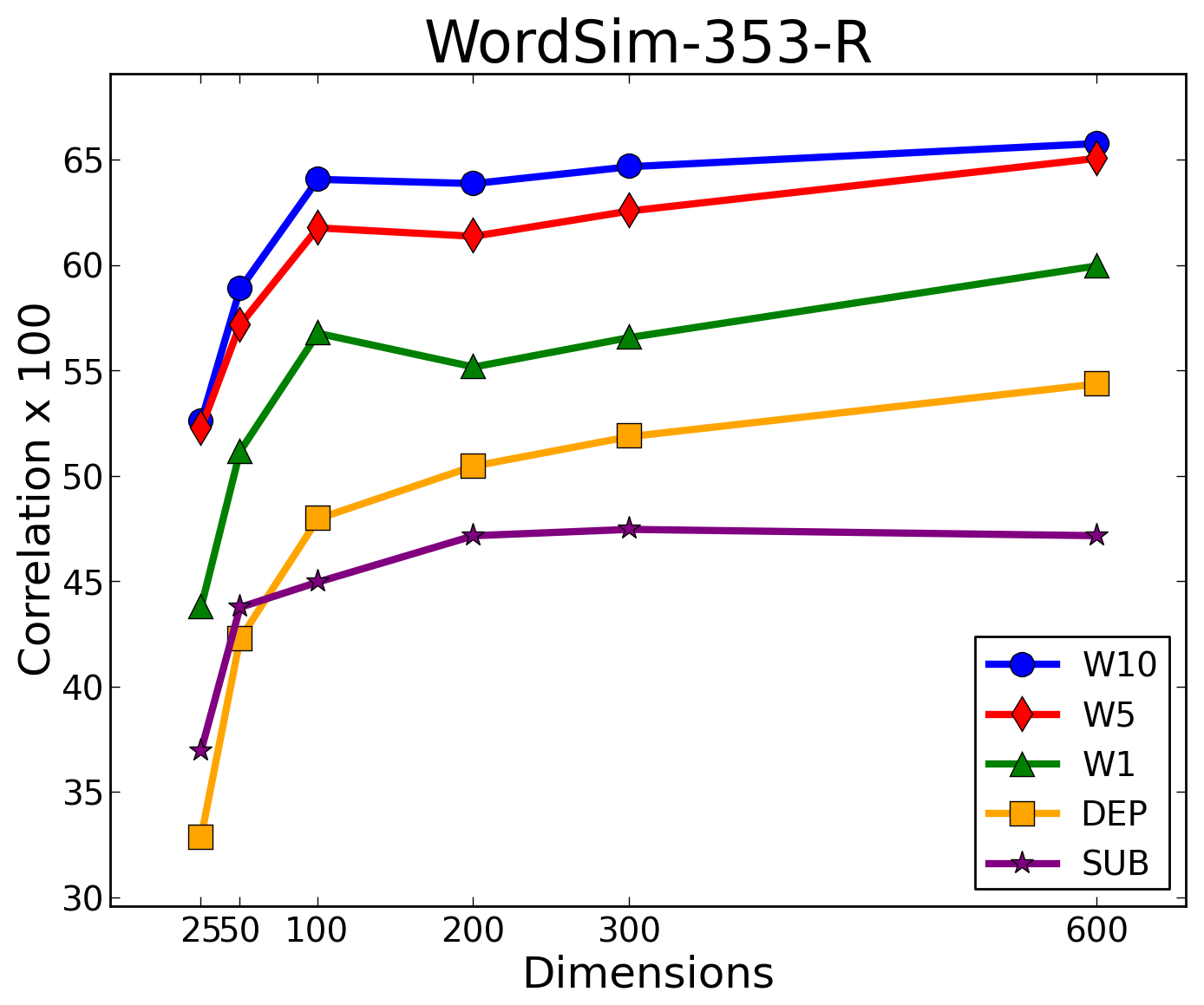} & \hspace{0.5cm}
\includegraphics[width=\figsize\textwidth]{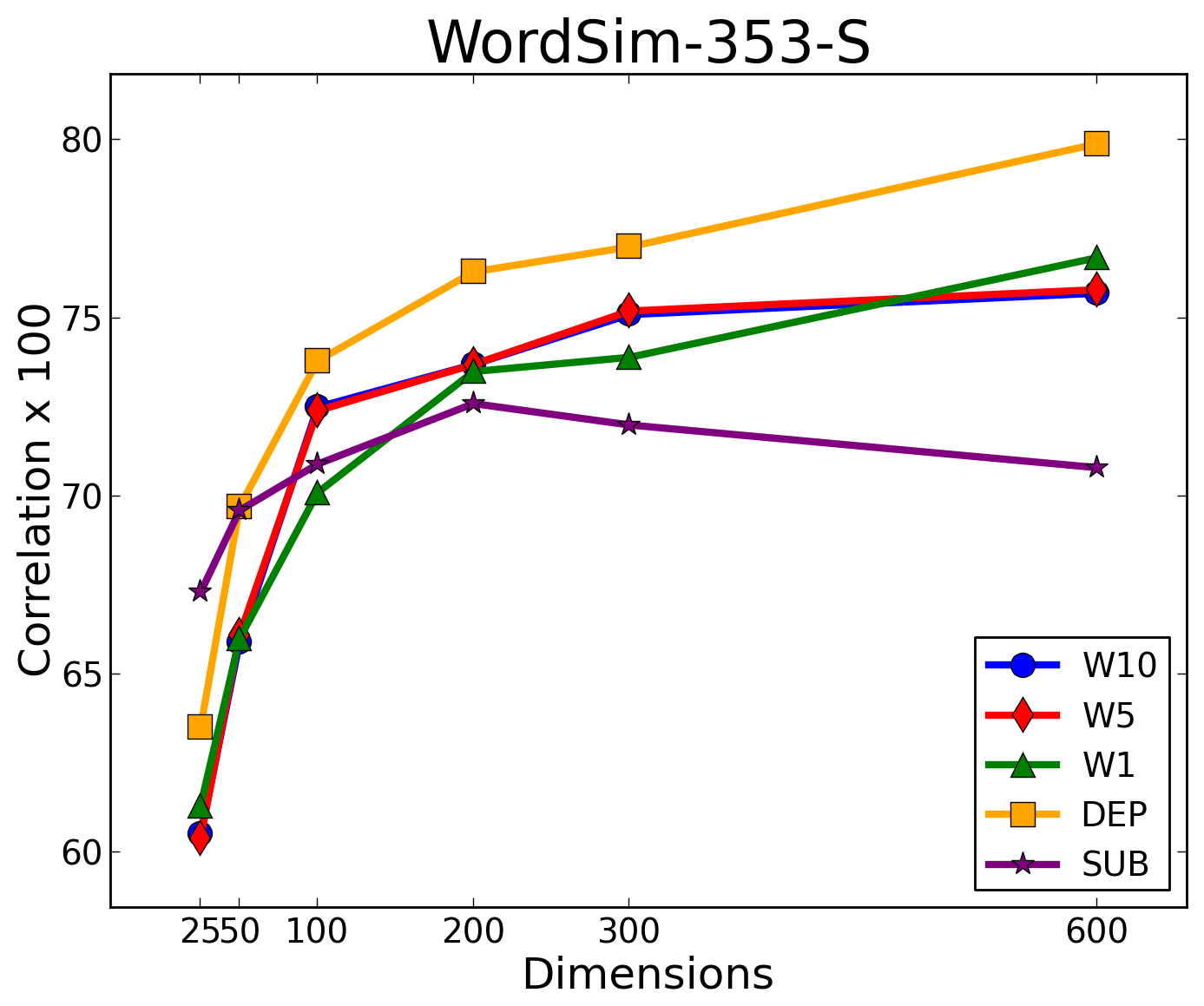}
\end{array}
$

$
\begin{array}{cc}
\includegraphics[width=\figsize\textwidth]{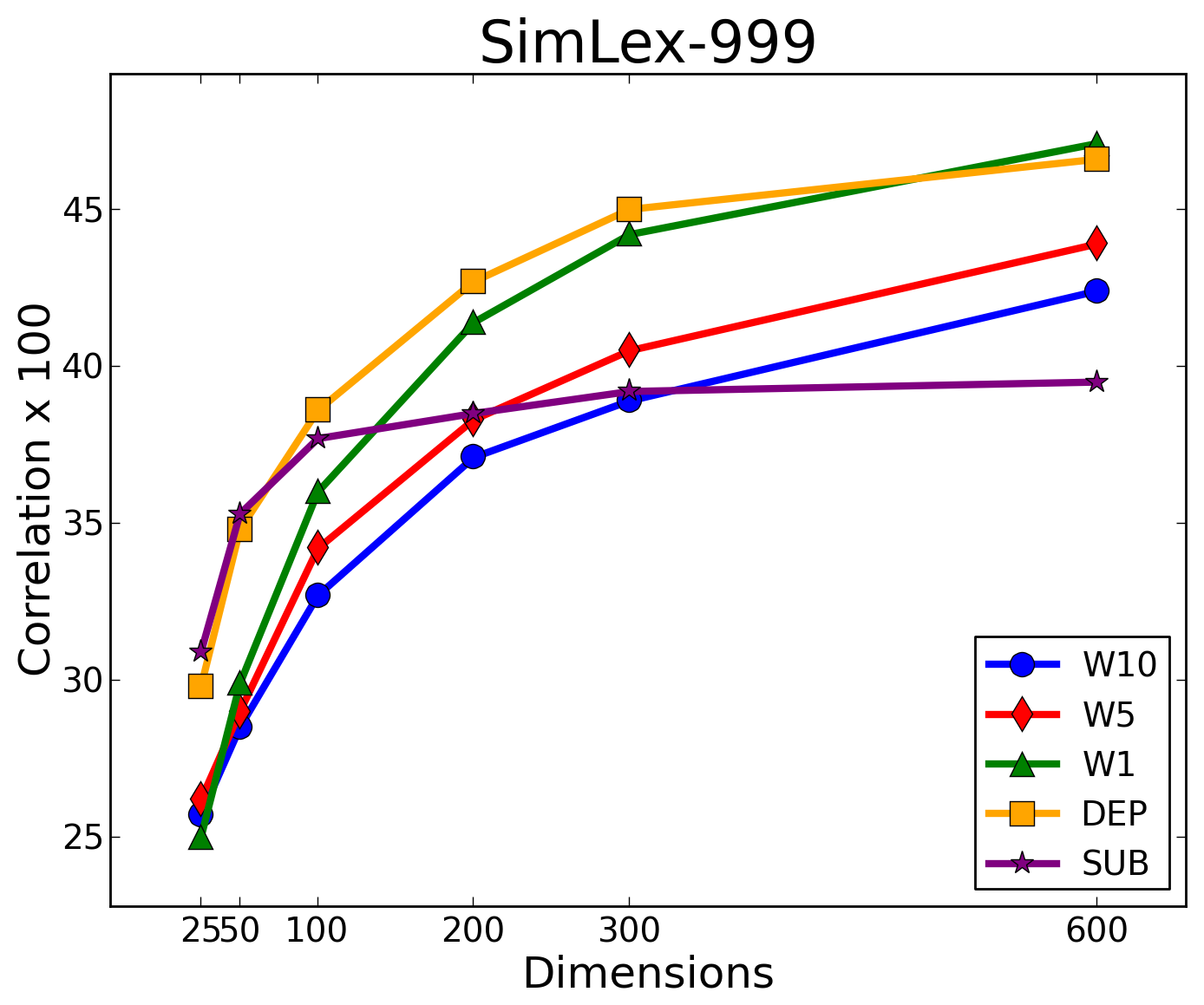} & \hspace{0.5cm}
\includegraphics[width=\figsize\textwidth]{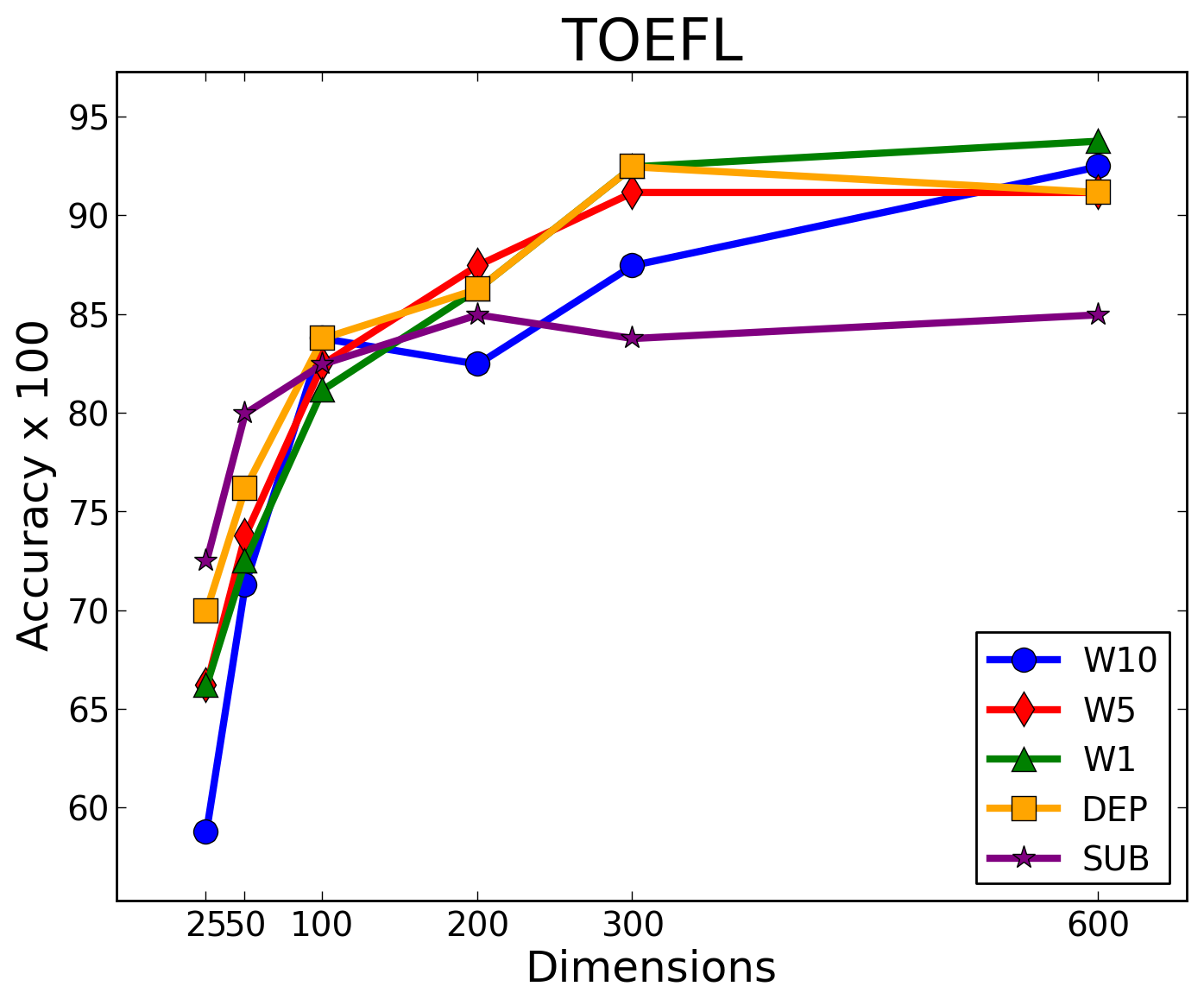}
\end{array}
$

\caption{Intrinsic tasks' results for embeddings learned with different types of contexts.
}
\label{fig:intrinsic_singles}
\end{figure*}

\paragraph{3) Coreference Resolution (\textsc{coref})} We used the Berkeley Coreference System \cite{durrett2013easy}, which achieves near state-of-the-art results with a log-linear supervised model. Most of the features in this model are associated with pairs of \emph{current} and \emph{antecedent} reference mentions, for which a coreference decision needs to be made. 
To evaluate the contribution of different word embedding types to this model, we extended it to support the following additional features: 
$\{a_{i}\}_{i=1..m}$, $\{c_{i}\}_{i=1..m}$ and $\{a_{i} \cdot c_{i}\}_{i=1..m}$, where $a_{i}$ or $c_{i}$ is the value of the $i$th dimension in a word embedding vector representing the antecedent or current mention, respectively. We considered two different word embedding representations for a mention: (1) the embedding of the head word of the mention and (2) the average embedding of all words in the mention. The features of both types of representations were presented to the learning model as inputs at the same time. 
They were added on top of Berkeley's full feature list (`FINAL') as described in \newcite{durrett2013easy}.
We evaluated our features on the CoNLL-2012 coreference shared task \cite{2012conll-coref}. 

\paragraph{4) Sentiment Analysis (\textsc{senti})} Following \newcite{faruqui-15}, we used a sentence-level binary decision version of the sentiment analysis task from \newcite{socher2013recursive}. In this setting, neutral sentences were discarded and all remaining sentences were labeled coarsely as positive or negative. Maintaining the original split into train/dev results, we get a dataset containing 6920/872 sentences.	
To evaluate different types of word embeddings, we represented each sentence as an average of its word embeddings and then used an L2-regularized logistic regression classifier trained on these features to predict the sentiment labels.

\begin{figure*}[t]
\centering

$
\begin{array}{cc}
\includegraphics[width=\figsize\textwidth]{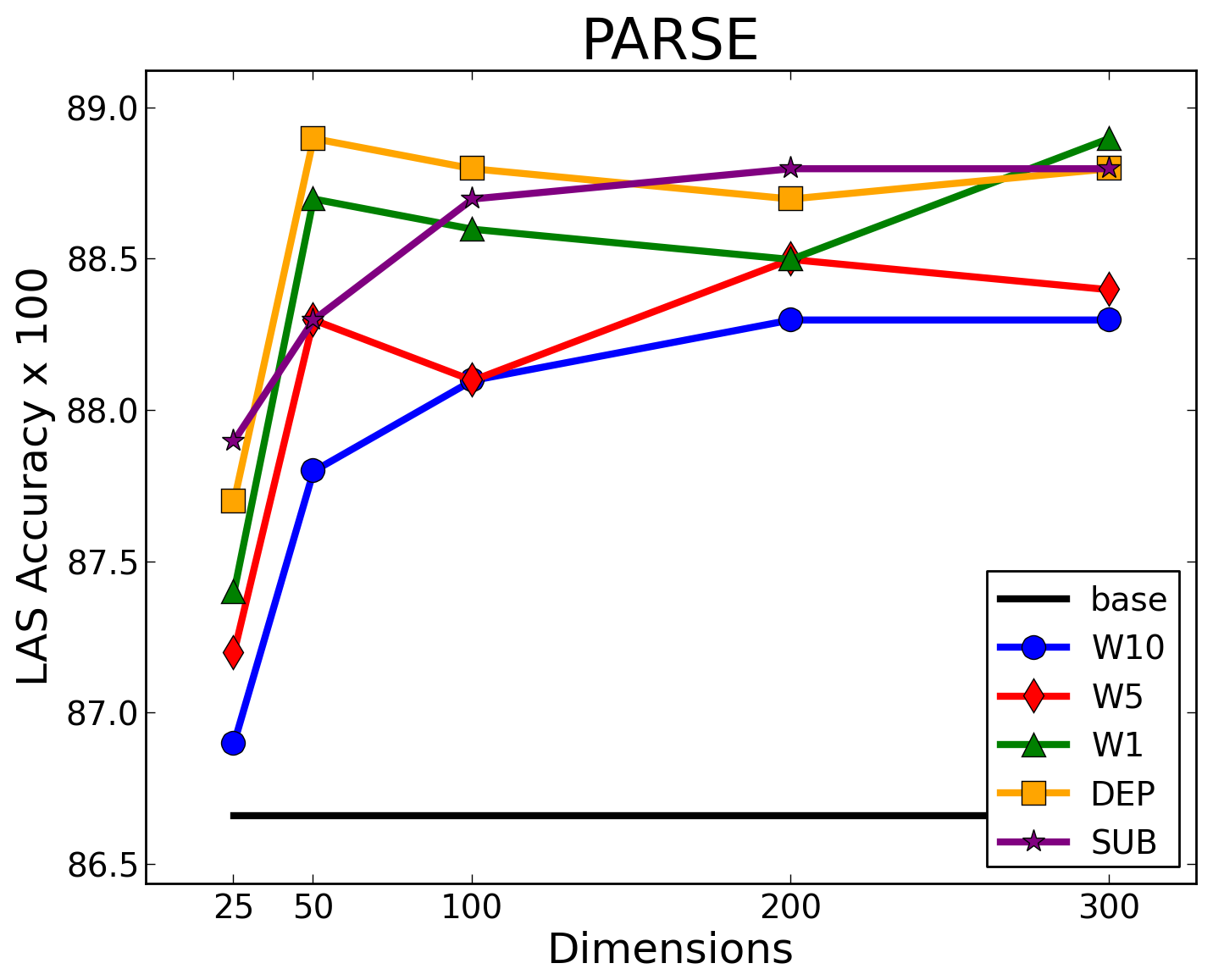} & \hspace{0.5cm}
\includegraphics[width=\figsize\textwidth]{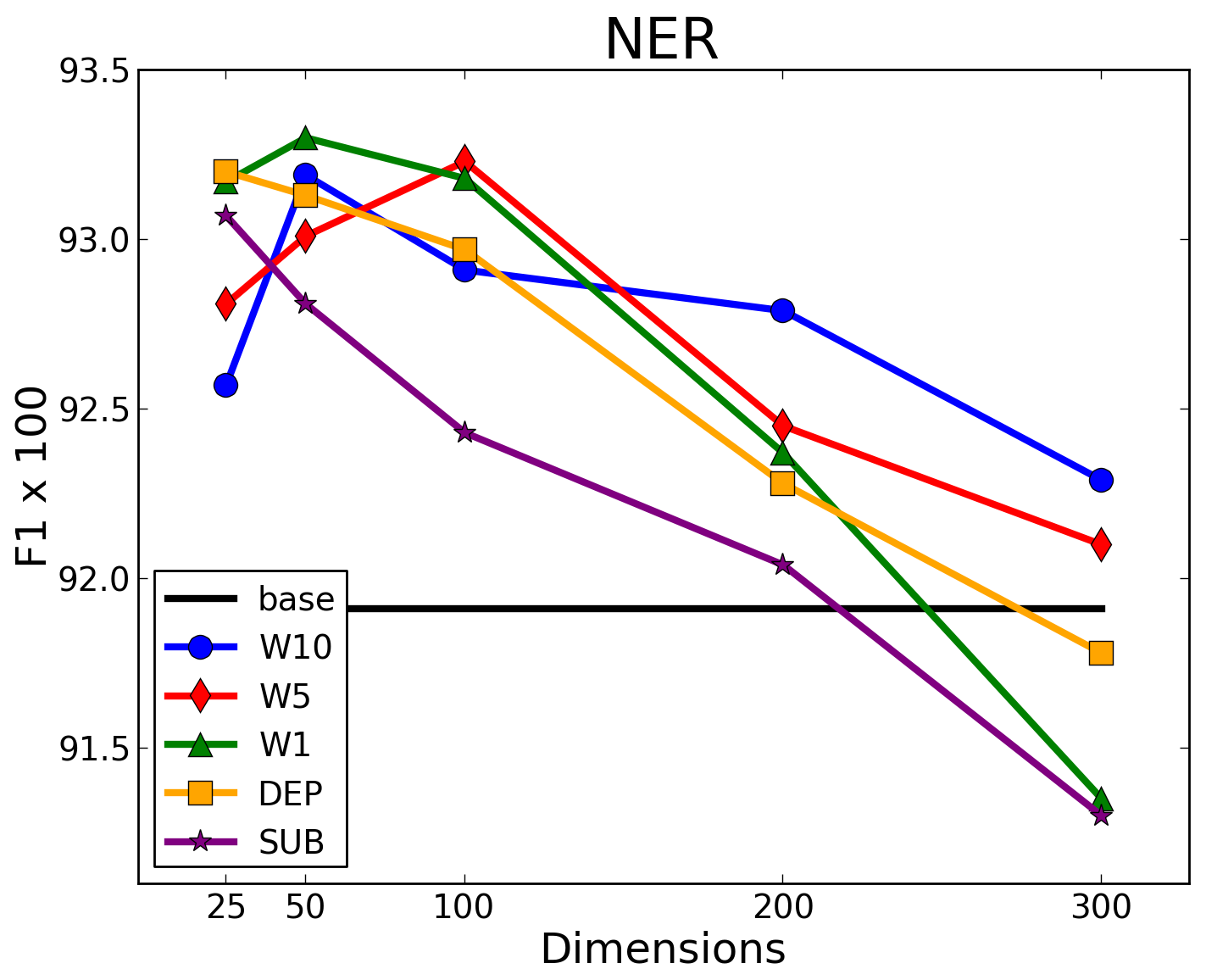}
\end{array}
$

$
\begin{array}{cc}
\includegraphics[width=\figsize\textwidth]{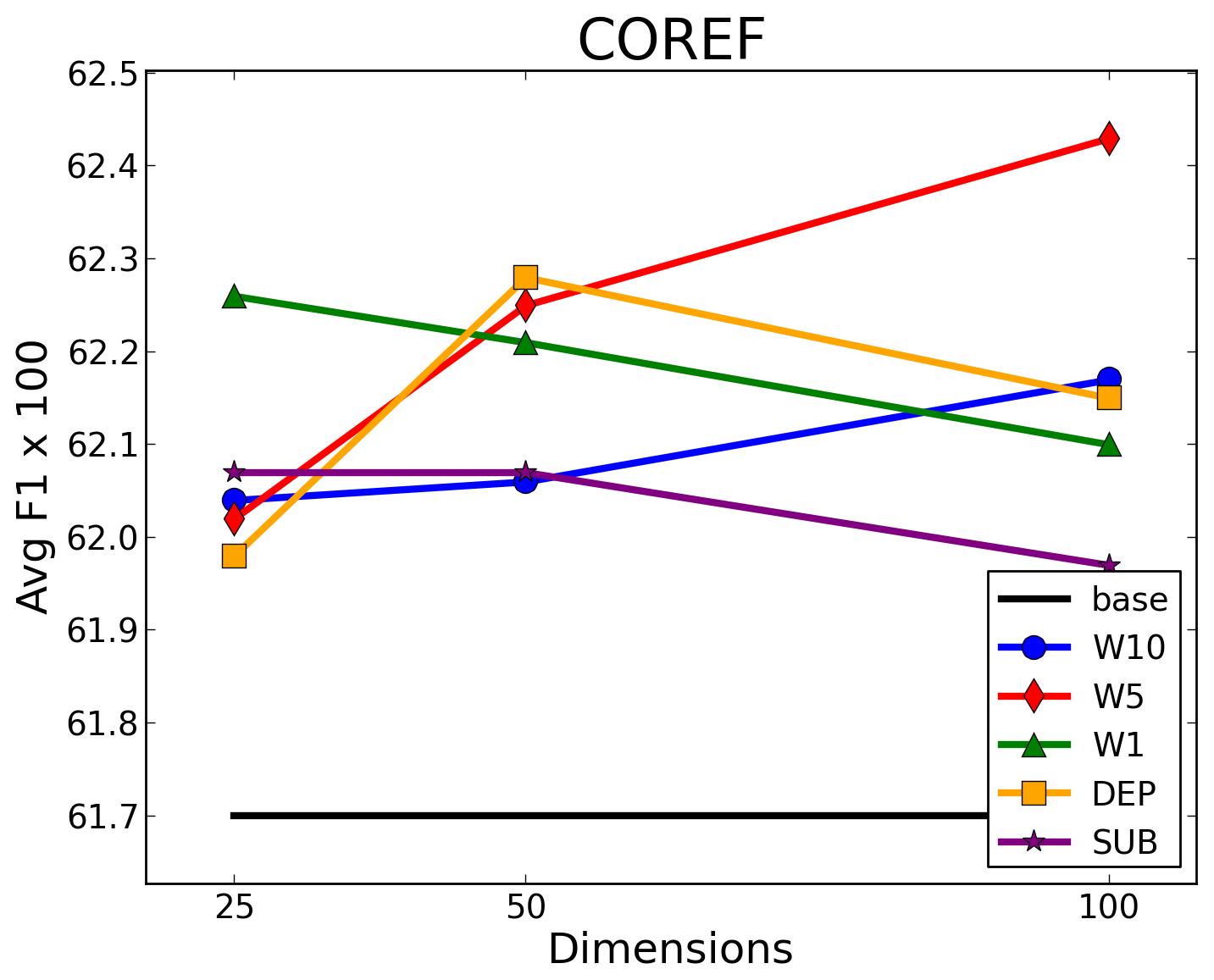} & \hspace{0.5cm}
\includegraphics[width=\figsize\textwidth]{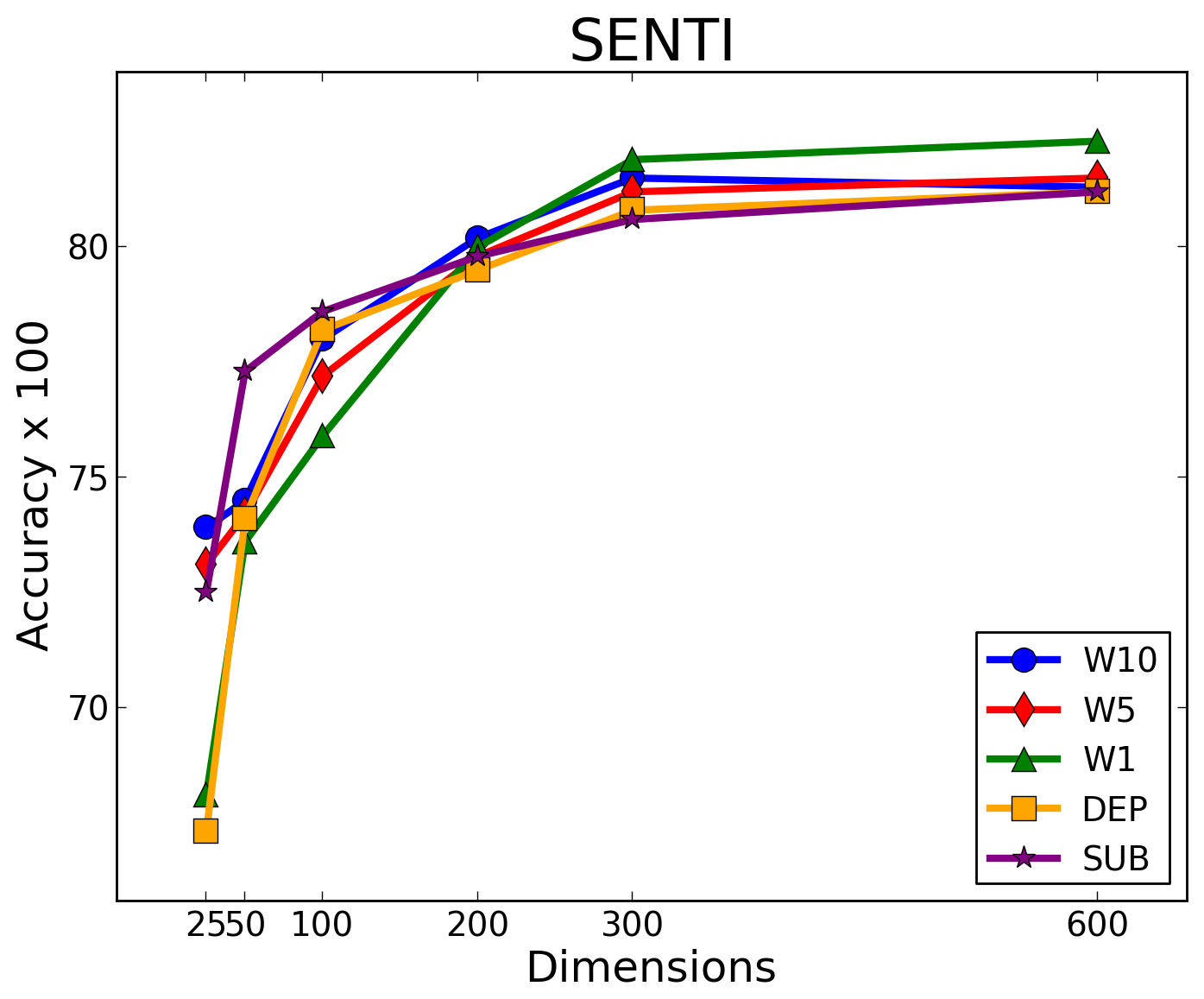}
\end{array}
$

\caption{Extrinsic tasks' development set results for embeddings learned with different types of contexts. `base' denotes the results with no word embedding features.
Due to computational limitations we tested \textsc{ner} and \textsc{parse} with only up to 300 dimensions embeddings, and \textsc{coref} with up to 100.
}
\label{fig:extrinsic_singles}
\end{figure*}

\section{Results}

\subsection{Intrinsic Results for Context Types}

The results on the intrinsic tasks
are illustrated in Figure \ref{fig:intrinsic_singles}. First, we see that the performance on all tasks generally increases with the number of dimensions, reaching near-optimal performance at around 300 dimensions, for all types of contexts. This is in line with similar observations on skip-gram word embeddings \cite{mikolov-13}.

Looking further, we observe that there are significant differences in the results when using different types of contexts. The effect of context choice is perhaps most evident in the WordSim-353-R task, which captures topical similarity.
As might be expected, in this benchmark, the largest-window word embeddings perform best. The performance decreases with the decrease in window size and then reaches significantly lower levels for dependency (DEP) and substitute-based (SUB) embeddings. 
Conversely, in WordSim-353-S and SimLex-999, both of which capture a more functional similarity, the DEP embeddings are the ones that perform best, strengthening similar observations in \newcite{levy2014dependencybased}. Finally, in the TOEFL benchmark, all contexts except for SUB, perform comparably.

\subsection{Extrinsic Results for Context Types}

The extrinsic tasks results are illustrated in Figure~\ref{fig:extrinsic_singles}. A first observation is that optimal extrinsic results may be reached with as few as 50 dimensions. Furthermore, performance may even degrade when using too many dimensions, as is most evident in the \textsc{ner} task.  This behavior presumably depends on various factors, such as the size of the labeled training data or the type of classifier used, and highlights the importance of tuning the dimensionality of word embeddings in extrinsic tasks. This is in contrast to intrinsic tasks, where higher dimensionality typically yields better results. 

Next, comparing the results of different types of contexts, we see, as might be expected, that dependency embeddings work best in the \textsc{parse} task. More generally, embeddings that do well in functional similarity intrinsic benchmarks and badly in topical ones (DEP, SUB and W1) work best for \textsc{parse}, while large window contexts perform worst, similar to observations in \newcite{bansal-14}.

In the rest of the tasks it's difficult to say which context works best for what.
One possible explanation to this in the case of \textsc{ner} and \textsc{coref} is that the embedding features are used as add-ons to an already competitive learning system. Therefore, the total improvement on top of a `no embedding' baseline is relatively small, leaving little room for significant differences between different contexts.

\begin{table}[t]
\centering
\begin{tabular}{| l| c |}
\hline
 Context type &  F1 x 100 \\
\hline
 DEP &    79.8 \\
 W1 &    79.3 \\
 SUB &   79.0 \\
 W10 &    78.1 \\
 W5 &    77.4 \\
 None &   71.8  \\
\hline
\end{tabular}
\caption{NER MUC out-of-domain results for different embeddings with dimensionality = 25.
}
\label{tab:ner_muc_singles}
\end{table}

We did find a more notable contribution of word embedding features to the overall system performance in the out-of-domain NER MUC evaluation, described in Table \ref{tab:ner_muc_singles}.
In this out-of-domain setting, all types of contexts achieve at least five points improvement over the baseline. Presumably, this is because continuous word embedding features are more robust to differences between train and test data, such as the typical vocabulary used. However, a detailed investigation of out-of-domain settings is out of scope for this paper and left for future work.

\subsection{Extrinsic Results for Combinations}

\begin{table*}[t]
\begin{center}
\small

\begin{tabular}{|l|l|l|l|l|l|}
  \hline
 Dimensions & Result & \multicolumn{1}{c|}{\textsc{senti}} & \multicolumn{1}{c|}{\textsc{parse}} & \multicolumn{1}{c|}{\textsc{ner}} & \multicolumn{1}{c|}{\textsc{coref}} \\
\hline
\multicolumn{1}{|c|}{\multirow{4}{*}{50}} & best+ & 74.3 (W10+W1) & 88.7 (W10+SUB) & \textbf{93.6} (W1+DEP) & \textbf{62.4} (W10+W1)\\
   & best & \textbf{77.3} (SUB) & \textbf{88.9} (W1) & 93.3 (W1) & 62.3 (DEP)\\
\cline{2-6}
 & mean+ & 72.7  & 88.3 & \textbf{93.3} & 62.1\\
   & mean & \textbf{74.7}  & \textbf{88.4} & 93.1 & \textbf{62.2}\\
   
   \hline
\multicolumn{1}{|c|}{\multirow{4}{*}{200}}    & best+ & \textbf{81.0} (W10+SUB) & \textbf{89.1} (W1+DEP) & \textbf{93.1} (W10+DEP) & \multicolumn{1}{c}{} \\
   & best & 80.2 (W10) & 88.8 (SUB) & 92.8 (W10) & \multicolumn{1}{c}{} \\
\cline{2-5}
    & mean+ & 79.1 & \textbf{88.9} & \textbf{92.8} &\multicolumn{1}{c}{} \\
   & mean & \textbf{79.9} & 88.6 & 92.4 &\multicolumn{1}{c}{} \\

\cline{1-5}
\multicolumn{1}{|c|}{\multirow{4}{*}{600}}    & best+ & \textbf{82.6} (W10+SUB)  &  \multicolumn{3}{c}{} \\
   & best & 82.3 (W1) & \multicolumn{3}{c}{} \\
\cline{2-3}
    & mean+ & \textbf{82.0}  & \multicolumn{3}{c}{} \\
   & mean & 81.5 & \multicolumn{3}{c}{} \\

\cline{1-3}
\end{tabular}

\caption{Extrinsic tasks development set results obtained with word embeddings concatenations. `best' and `best+' are the best results achieved across all single context types and context concatenations, respectively (best performing embedding indicated in parenthesis). `mean' and `mean+' are the mean results for the same. Due to computational limitations of the employed systems, some of the evaluations were not performed.
}
\label{tab:concats}
\end{center}
\end{table*}

\begin{figure}[t]
\centering

$
\includegraphics[width=\figsize\textwidth]{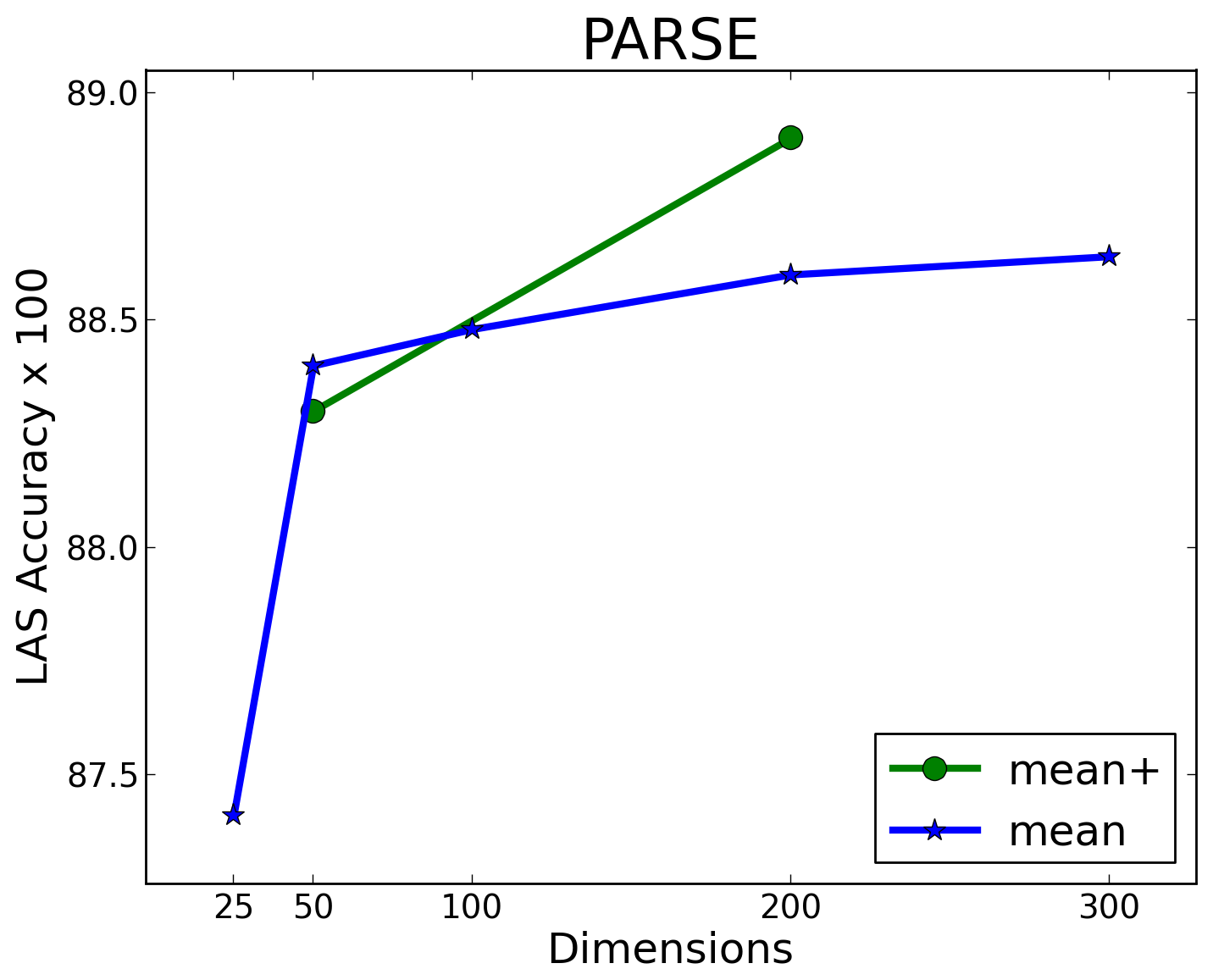} 
$

$
\includegraphics[width=\figsize\textwidth]{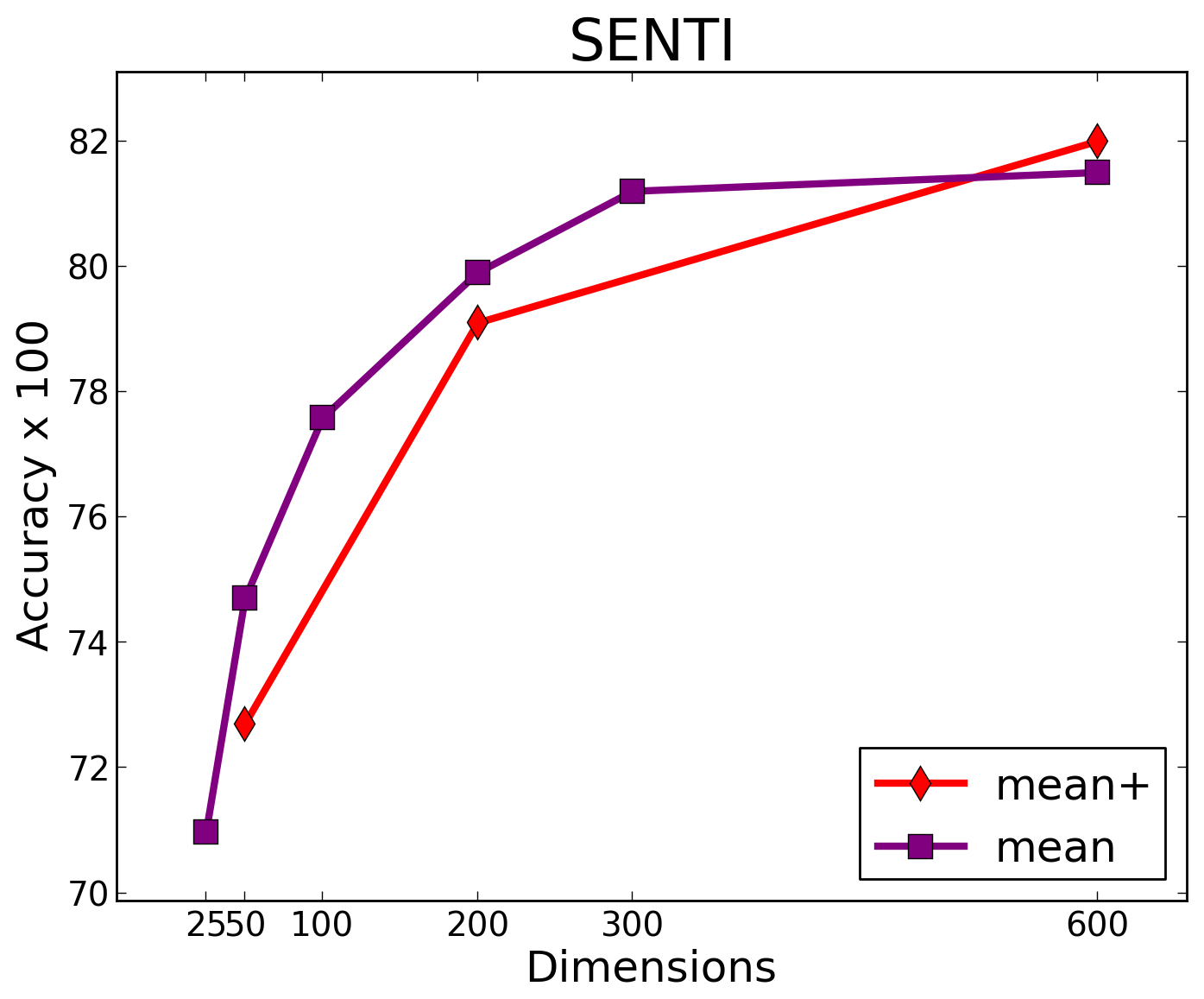} 
$

\caption{Mean development set results for the tasks \textsc{parse} and \textsc{senti}. `mean' and 'mean+' stand for mean results across all single context types and context concatenations, respectively.
}
\label{fig:means}
\end{figure}

A comparison of the results obtained on the extrinsic tasks using the word embedding concatenations (\emph{concats}), described in section \ref{subsec:concats}, versus the original single context word embeddings (\emph{singles}), appears in Table \ref{tab:concats}. To control for dimensionality, concats are always compared against singles with identical dimensionality. For example, the 200-dimensional concat W10+DEP$^{200}$, which is a concatenation of W10$^{100}$ and DEP$^{100}$, is compared against 200-dimensional singles, such as W10$^{200}$.

Looking at the results, it seems like the benefit from concatenation depends on the dimensionality and task at hand, as also illustrated in Figure~\ref{fig:means}. Given task $X$ and dimensionality $d$, if $\frac{d}{2}$ is in the range where increasing the dimensionality yields significant improvement on task $X$,
then it's better to simply increase dimensionality of singles from $\frac{d}{2}$ to $d$ rather than concatenate. The most evident example for this are the results on the \textsc{senti} task with $d=50$. In this case, the benefit from concatenating two 25-dimensional singles is notably lower than that of using a single 50-dimensional word embedding. On the other hand, if  $\frac{d}{2}$ is in the range where near-optimal performance is reached on task $X$, then concatenation seems to pay off. This can be seen in \textsc{senti} with $d=600$, \textsc{parse} with $d=200$, and \textsc{ner} with $d=50$. 
More concretely, looking at the best performing concatenations, it seems like combinations of the topical W10 embedding with one of the more functional ones, SUB, DEP or W1, typically perform best, suggesting that there is added value in combining embeddings of different nature.

Finally, our experiments with the methods using SVD (section \ref{subsec:svd}) and CCA  (section \ref{subsec:cca}) yielded degraded performance compared to single word embeddings for all extrinsic tasks and therefore are not reported for brevity. 
These results seem to further strengthen the hypothesis that the information captured with varied types of context is different and complementary, and therefore it is beneficial to preserve these differences as in our concatenation approach.

\section{Related Work}
\label{sec:related}

There are a number of recent works whose goal is a broad evaluation of the performance of different word embeddings on a range of tasks. However, to the best of our knowledge, none of them focus on embeddings learned with diverse context types as we do.
\newcite{levy2015improving}, \newcite{lapesa2014large}, and \newcite{lai2015generate} evaluate several  design choices when learning word representations. However, \newcite{levy2015improving} and \newcite{lapesa2014large} perform only intrinsic evaluations and restrict context representation to word windows, while \newcite{lai2015generate} do perform extrinsic evaluations, but restrict their context representation to a word window with the default size of~5.
\newcite{schnabel2015evaluation} and \newcite{lingevaluation2015} report low correlation between intrinsic and extrinsic results with different word embeddings (they did not evaluate different context types), which is consistent with differences we found between intrinsic and extrinsic performance patterns in all tasks, except parsing. \newcite{bansal-14} show that functional  (dependency-based and small-window) embeddings  yield higher parsing improvements than topical (large-window) embeddings, which is consistent with our findings.

Several works focus on particular types of contexts for learning word embeddings. 
\newcite{cirik2014substitute} investigates  S-CODE word embeddings based on substitute word contexts. \newcite{wang-15} and \newcite{yulianot2015} propose extensions to the standard window-based context modeling.
Alternatively, another recent popular line of work \cite{faruqui-15,kielaspecializing2015} attempts to improve word embeddings by using manually-constructed resources, such as WordNet. These techniques could be complementary to our work.
Finally, \newcite{yin2015learning} and \newcite{goikoetxea2016single} propose word embeddings combinations, using methods such as concatenation and CCA, but evaluate mostly on intrinsic tasks and do not consider different types of contexts.

\section{Conclusions}
In this paper we evaluated skip-gram word embeddings on multiple intrinsic and extrinsic NLP tasks, varying dimensionality and type of context.
We show that while the best practices for setting skip-gram hyperparameters typically yield good results on intrinsic tasks, success on extrinsic tasks requires more careful thought.
Specifically, we suggest that picking the optimal dimensionality and context type are critical for obtaining the best accuracy on extrinsic tasks and are typically task-specific.
Further improvements can often be achieved by combining complementary word embeddings of different context types with the right dimensionality.

\section*{Acknowledgments}
We thank Do Kook Choe for providing us the jackknifed version of WSJ. We also wish to thank the IBM Watson team for helpful discussions and our anonymous reviewers for their comments.
This work was partially supported by the Israel
Science Foundation grant 880/12 and the German
Research Foundation through the German-Israeli
Project Cooperation (DIP, grant DA 1600/1-1).

\bibliography{naaclhlt2016}
\bibliographystyle{naaclhlt2016}

\end{document}